\documentclass[letterpaper, 10 pt, conference]{ieeeconf}  
\usepackage[utf8]{inputenc}
\usepackage{graphicx}
\usepackage{amsmath,amssymb,amsfonts}
\usepackage{subcaption}
\usepackage[colorlinks=true, linkcolor=blue, citecolor=blue,anchorcolor=blue, urlcolor=blue]{hyperref}
\usepackage{algorithmic}
\usepackage{cite}
\usepackage{textcomp}
\usepackage{xcolor}
\usepackage{longtable}
\usepackage{mathtools}
\usepackage{multirow} 

\IEEEoverridecommandlockouts                              

 \title{\LARGE \bf
 	The Stiffness of 3-PRS PM Across Parasitic and Orientational Workspace
 }

 \author{Hassen Nigatu$^{1,2}$, Li Jihao$^{1,2}$, Keqi Zhu$^ {1}$, Junhan Zhang$^ {1}$, Haotian Guo$^{1}$, Guodong Lu $^{1,3}$, and Doik Kim$^{*4}$
 	\thanks{$^{1}$ Robot Perception and Grasp Lab, School of Mechanical Engineering, Zhejiang University, Hangzhou 310027, China.
 		{\tt\small }}%
 	\thanks{$^{2}$ Robotics Research Center of Yuyao (Robotics Institute of Zhejiang University), Yuyao Technology Innovation Center, No.479 Ye Shan Road , Yuyao, Ningbo Shi, Zhejiang Province, China.
 		{\tt\small }}%
 	\thanks{$^{3}$ The State Key Laboratory of Fluid Power and Mechatronic Systems, School of Mechanical Engineering, and the Engineering Research Center for Design Engineering and Digital Twin of Zhejiang Province, School of Mechanical Engineering, Zhejiang University, Hangzhou 310027, China.
 		{\tt\small doikkim@kist.re.kr, hassity@gmail.com}}%
 	\thanks{$^{4}$ xR Lab, Center for Intelligent and Interactive Robotics, Artificial Intelligence and Robot Institute, Korea Institute of Science and Technology (KIST), South Korea.
 		{\tt\small doikkim@kist.re.kr, hassity@gmail.com}}%
 	\thanks{Correspondence: Doik Kim, {\tt\small doikkim@kist.re.kr}}
 }

\begin{document}

 \newcommand{\join}{\curlyvee}
 \newcommand{\meet}{\curlywedge}

\maketitle

\begin{abstract}	
	This study investigates the stiffness characteristics of the Sprint Z3 head, also known as 3-PRS Parallel Kinematics Machines, which are among the most extensively researched and viably successful manipulators for precision machining applications. Despite the wealth of research on these robotic manipulators, no previous work has demonstrated their stiffness performance within the parasitic motion space. Such an undesired motion influences their stiffness properties, as stiffness is configuration-dependent. Addressing this gap, this paper develops a stiffness model that accounts for both the velocity-level parasitic motion space and the regular workspace. Numerical simulations are provided to illustrate the stiffness characteristics of the manipulator across all considered spaces. The results indicate that the stiffness profile within the parasitic motion space is both shallower and the values are smaller when compared to the stiffness distribution across the orientation workspace. This implies that evaluating a manipulator's performance adequately requires assessing its ability to resist external loads during parasitic motion. Therefore, comprehending this aspect is crucial for redesigning components to enhance overall stiffness.  
\end{abstract}

\section{Introduction}
Parasitic motion is a well-known drawback of lower degrees of freedom (DoFs) parallel manipulators (PMs) \cite{Carretero2000,Nigatu2021_mmt}. Sprint Z3, also known as 3-PRS PM is one of the successful manipulator used in high precision machining in aerospace and automobile industries \cite{wahl2002}.  This unusual phenomenon occurs at the center of the moving platform along and about the constrained directions. As long as this undesired motion is not eliminated from the workspace via optimization \cite{Carretero2000,Nigatu2021,Nigatu2021_mmt}, the authors believe that the performance of the manipulator within this space must be explicitly considered. It is evident that stiffness is one of the most important performance specifications for parallel robots, especially those used for machine tools \cite{Huang2002}. Consequently, there has been significant research in this area \cite{Hoevenaars2016,Li2010,Liu2017,Tang2017}. However, most studies focusing on this key property are dedicated to evaluating its distribution across the permissible workspace \cite{Li2010,Liu2017,Li2013,Zhao2016,Sun2009}, among numerous other studies. Addressing this gap, our paper evaluates the stiffness distribution of the manipulator over the spaces spanned by the parasitic and independent task space variables. 

In this study, we begin by formulating the analytic inverse Jacobian \cite{doik,Kim2002,Nigatu2021_mmt}, followed by the development of the parasitic motion equation at the velocity level, in accordance with the velocity model. Subsequently, we derive the stiffness equation utilizing the inverse Jacobian, principles of virtual work, Hooke's law, and the material and geometric properties of the links and joints.

The structure of this paper is organized as follows: Section \ref{sec: discription} introduces the manipulator's geometric and kinematic characteristics. The inverse Jacobian of the manipulator is detailed in Section \ref{sec: velocity_equation}, and the stiffness model is elucidated in Section \ref{sec: stiffness_formulation}. Section \ref{sec: matrix_k} derives the stiffness matrix, conceptualizing each limb's components as a series-connected spring system, with numerical simulations discussed in Section \ref{sec:numerical_simulation}. The paper concludes in Section \ref{sec:conclusions}.

\section{Kinematic description of Sprint Z3}  \label{sec: discription}

\begin{figure}[htb!]
	\centering
	\includegraphics[width=\columnwidth]{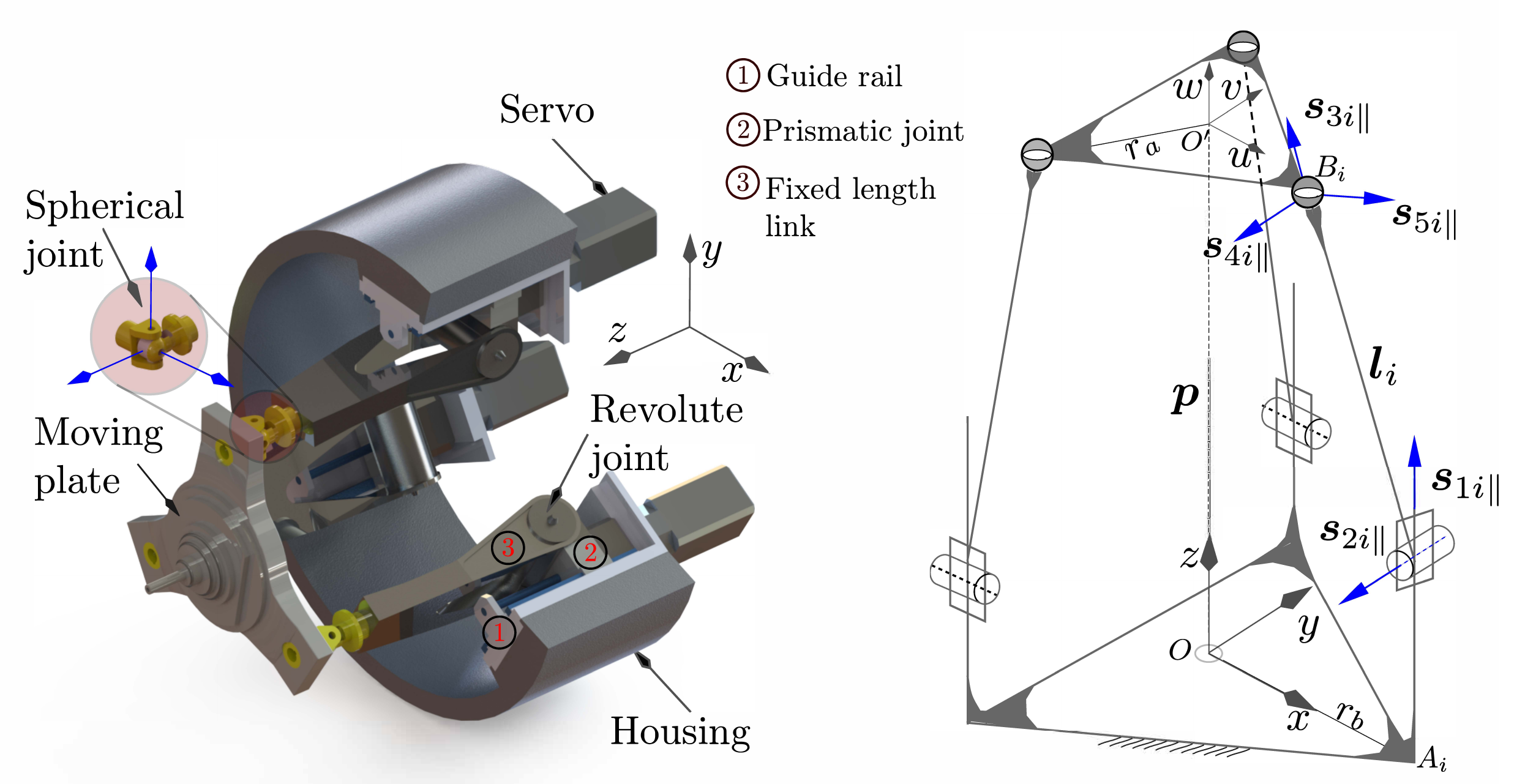} 
	\caption{Sprint Z3 parallel manipulator: CAD model (left) Schematic (right)}
	\label{fig:spritnz3}
\end{figure}

As illustrated in Fig. (\ref{fig:spritnz3}), the Sprint Z3 parallel manipulator is comprised of a moving platform, limbs with Prismatic-Revolute-Spherical (PRS) joint sequences, and a base platform. The position vectors $\boldsymbol{l}_i$ and the directional vectors $\boldsymbol{s}_{ij\parallel}$ corresponding to the fixed-length links and the joints of the $i^{th}$ limbs, respectively.     

Servo-driven prismatic joints enable the limbs to extend or retract along the z-axis, guided by rails affixed to the housing. The revolute joints, which are passive, allow the limbs to rotate parallel to the base platform, while the spherical joints at the limb's terminus connect to the moving platform, endowing it with a full range of spatial orientation. The radii of the moving platform and base are denoted by the scalars $r_a$ and $r_b$, respectively. The vector $\boldsymbol{p}$ represents the position of the moving platform relative to the base. This configuration restricts the moving platform to translational movement along the z-axis and permits tip and tilt motions about the x and y axes of the base frame.

\section{Velocity Level Parasitic Motion } \label{sec: velocity_equation}

In most cases, the control and motion equation of these manipulators are formulated at the velocity level. This is due to the fact that embedding constraints is convenient and more intuitive with instantaneous kinematics at the velocity level, allowing for straightforward integration of the manipulator's motion constraints \cite{doik,Kim2002}. Hence, the parasitic motion equation presented in this paper is derived solely from the inverse Jacobian matrix and high precision numerical integration is performed to get the pose.

The analytic inverse velocity equation of the Sprint Z3 parallel robot is given as: 

\begin{equation}
	\begin{aligned}
		\boldsymbol{\dot{q}} &=  \boldsymbol{G}^\top \boldsymbol{\mathcal{\dot{X}}} \\ 
		\begin{bmatrix}
			\dot{\boldsymbol{q}}_a \\ \dot{\boldsymbol{q}}_c = \boldsymbol{0}
		\end{bmatrix} &= 
		\begin{bmatrix}
			\boldsymbol{G}_a \\  \boldsymbol{G}_c
		\end{bmatrix} \begin{bmatrix}  \boldsymbol{v} \\ \boldsymbol{\omega} \end{bmatrix}\\ 		
		 &=
		\begin{bmatrix}
			\boldsymbol{G}_{av}^\top & \boldsymbol{G}_{a\omega}^\top\\ \boldsymbol{G}_{cv}^\top & \boldsymbol{G}_{c\omega}^\top 
		\end{bmatrix} \begin{bmatrix}  \boldsymbol{v} \\ \boldsymbol{\omega} \end{bmatrix}
		\label{eq:irk}
	\end{aligned}
\end{equation}
where $\boldsymbol{G}$ is the inverse Jacobian partitioned into sub-matrices corresponding to motion and constraints, $\boldsymbol{\dot{q}} $ is the joint rate,  $\boldsymbol{\mathcal{\dot{X}}}$  is the constraint compatible task motion which comprises both the parasitic and independent velocity components.

Here, the detailed entries of the matrix are further computed using the analytic reciprocal screw approach \cite{doik,Nigatu2021,Nigatu2021_mmt}.

\begin{equation}
	\boldsymbol{G}^\top=
	\begin{bmatrix}
		\dfrac{\boldsymbol{l}_{i}^\top}{{\boldsymbol{l}_{i}^\top s_{1i\parallel}}} & \dfrac{(\boldsymbol{l}_{i}\times \boldsymbol{a}_i)^\top}{{{\boldsymbol{l}_{i}^\top s_{1i\parallel}}}}  \\
		\boldsymbol{s}_{2i{\parallel}}^\top & (\boldsymbol{s}_{2i{\parallel}}\times \boldsymbol{a}_i)^\top \\
	\end{bmatrix}  ~\text{for} ~ i = 1,2,3 \label{eq:Gprs}
\end{equation}

where $\boldsymbol{l}_{i}$ is a vector extended from the center of revolute joints to the center of spherical joints and $\boldsymbol{a}_i$ is a position vector extended from $O'$ to the center of the spherical joints. 
 
In Eq. (\ref{eq:irk}), to ensure that $\boldsymbol{\mathcal{\dot{X}}}$ is constraint compatible, the user desired motion $\boldsymbol{\mathcal{\dot{X}}}_a$ can be transformed into $\boldsymbol{\mathcal{\dot{X}}}$  using the constraint matrix $\boldsymbol{G}_{c}^\top = \begin{bmatrix} \boldsymbol{G}_{cv}^\top & \boldsymbol{G}_{c\omega}^\top  \end{bmatrix}  $ as:

\begin{equation}
	\begin{aligned}
		\boldsymbol{G}_{c}^\top \boldsymbol{\mathcal{\dot{X}}} &= \boldsymbol{0}  \\
		\boldsymbol{G}_{c}^\top ( \boldsymbol{I} -\boldsymbol{G}_{c}\boldsymbol{G}_{c}^+) \boldsymbol{\mathcal{\dot{X}}}_a &= \boldsymbol{0}
	\end{aligned} \label{eq:projection}
\end{equation} 
where $\boldsymbol{I} -\boldsymbol{G}_{c}\boldsymbol{G}_{c}^+$ in Eq. (\ref{eq:projection}) projects $\boldsymbol{\mathcal{\dot{X}}}_a$ into the motion space of the manipulator, and $\boldsymbol{G}_{c}^+$ is the pseudo-inverse. Otherwise, if no filter via projection is applied, $\boldsymbol{\mathcal{\dot{X}}}_a$ without parasitic motion could lead to inaccuracies or catastrophic consequences, such as structural breakage.

From the relations given in Eq. (\ref{eq:Gprs}) and (\ref{eq:projection}), and noting that $\boldsymbol{G}_{ci}^\top = \begin{bmatrix}  \boldsymbol{s}_{2i{\parallel}}^\top & (\boldsymbol{s}_{2i{\parallel}}\times \boldsymbol{a}_i)^\top \end{bmatrix}$,  the following relation can be obtained

\begin{equation}
	\begin{aligned}
		\boldsymbol{G}_{c}^\top \boldsymbol{\mathcal{\dot{X}}} &= \boldsymbol{0}  \\
		\begin{bmatrix}  \boldsymbol{G}_{cv}^\top & \boldsymbol{G}_{c\omega}^\top \end{bmatrix} \begin{bmatrix}  \boldsymbol{v} \\ \boldsymbol{\omega} \end{bmatrix} &= \boldsymbol{0}  \\
		\begin{bmatrix}  \boldsymbol{s}_{21\parallel}^\top & (\boldsymbol{s}_{21\parallel}\times \boldsymbol{a}_1)^\top \\   \boldsymbol{s}_{22\parallel}^\top & (\boldsymbol{s}_{22\parallel}\times \boldsymbol{a}_2)^\top \\  \boldsymbol{s}_{23\parallel}^\top & (\boldsymbol{s}_{23\parallel}\times \boldsymbol{a}_3)^\top \end{bmatrix} \begin{bmatrix}  \boldsymbol{v}_x \\\boldsymbol{v}_y \\\boldsymbol{v}_z   \\ \boldsymbol{\omega}_x \\ \boldsymbol{\omega}_y \\ \boldsymbol{\omega}_z \end{bmatrix}  &= \boldsymbol{0}
	\end{aligned}
\end{equation}  

given $ \boldsymbol{s}_{2i\parallel} = \boldsymbol{R}_z(\xi_i)\boldsymbol{T}_y(1) =  \begin{bmatrix}  -s\xi_i & c\xi_i & 0\end{bmatrix}^\top $ while $\xi_i = (i-1)\dfrac{2\pi}{3}$ and $\boldsymbol{T}_y(1) = \begin{bmatrix} 0 &1 &0  \end{bmatrix} ^\top  $.

Taking the $i^{th}$ row of $\boldsymbol{G}_{c}^\top$ and multiplying it with the corresponding element of $\boldsymbol{\mathcal{\dot{X}}}$, we can get the following relation. 

\begin{equation}
	\begin{aligned}
		-v_x s\xi_i + v_y c\xi_i + a_{iz}c\xi_i \omega_x + a_{iz}s\xi_i \omega_y + \\ ( -a_{ix}c\xi_i - a_{iy}s\xi_i)\omega_z = 0 
	\end{aligned} \label{eq:con_ith_exp}
\end{equation}

In the Sprint Z3 robot, we know that $v_x$, $v_y$ and $\omega_z$  are parasitic terms while $\omega_x$, $\omega_y$ and $v_z$ are independent components spanning the motion space. Hence, collecting $v_x$, $v_y$ and $\omega_z$ terms and their coefficients in Eq. (\ref{eq:con_ith_exp}) to one side leads us to: 

\begin{equation}
	\begin{aligned}
		\begin{bmatrix} -s\xi_i & c\xi_i (-a_{ix} c\xi_i - a_{iy} s\xi_i ) \end{bmatrix} \begin{bmatrix} v_x \\ v_y \\ \omega_z\end{bmatrix} =  \\  \begin{bmatrix}  -a_{iz}c\xi_i & -a_{iz} s\xi_i \end{bmatrix}  \begin{bmatrix} \omega_x \\ \omega_y \end{bmatrix} \\ 
	\end{aligned} \label{eq:con_ith_collected}
\end{equation}

Writing Eq. (\ref{eq:con_ith_collected}) for all three limbs:

\begin{equation}
	\begin{aligned}
		\boldsymbol{C}_1 \dot{\boldsymbol{x}}_d = \boldsymbol{C}_2 \dot{\boldsymbol{x}}_i \\ 
	\end{aligned} \label{eq:coupling}
\end{equation}

where $\boldsymbol{C}_1 = \begin{bmatrix} -s\xi_1 & c\xi_1 (-a_{1x} c\xi_1 - a_{1y} s\xi_1 ) \\  -s\xi_2 & c\xi_2 (-a_{2x} c\xi_2 - a_{2y} s\xi_2)  \\ -s\xi_3 & c\xi_3 (-a_{3x} c\xi_3 - a_{3y} s\xi_3 ) \end{bmatrix}$, $\dot{\boldsymbol{x}}_d = \begin{bmatrix} v_x \\ v_y \\ \omega_z\end{bmatrix} $, $\boldsymbol{C}_2 = \begin{bmatrix}  -a_{1z}c\xi_1 & -a_{1z} s\xi_1 \\  -a_{2z}c\xi_2 & -a_{2z} s\xi_2 \\ -a_{3z}c\xi_3 & -a_{3z} s\xi_3 \end{bmatrix} $ and $ \dot{\boldsymbol{x}}_i =  \begin{bmatrix} \omega_x \\ \omega_y \end{bmatrix}$ 

Finally, by inverting $\boldsymbol{C}_1$ we get the coupling relation between parasitic and independent motion as shown below.

\begin{equation}
	\begin{aligned}
		\dot{\boldsymbol{x}}_d &= \boldsymbol{C}_1^{-1}\boldsymbol{C}_2 \dot{\boldsymbol{x}}_i  \\ 
		\begin{bmatrix} v_x \\ v_y \\ \omega_z\end{bmatrix} &= \begin{bmatrix} \boldsymbol{C}_1^{-1}\boldsymbol{C}_2 & \boldsymbol{0}  \end{bmatrix}  \begin{bmatrix} \omega_x \\ \omega_y  \\ v_z\end{bmatrix} \\ 
	\end{aligned} \label{eq:coupling_1}
\end{equation}

From Eq. (\ref{eq:coupling_1}), namely the relationship between independent and parasitic task space motion, we clearly observe that $v_z$ has no contribution in generating of the parasitic motion. It is also evident that the parasitic motion is influenced by the structural constraint embedded in the Jacobian matrix.

\section{Parasitic Motion and the Stiffness Relation} \label{sec:Paramo_Stiff}

We knew that the parasitic motion terms are $v_x,v_y$ and $\omega_z$. Thus, collecting the parasitic terms together, Eq. (\ref{eq:irk}) and Eq. (\ref{eq:coupling_1}) can be reorganized as 

\begin{equation}
		\begin{bmatrix} \boldsymbol{\dot{q}}_a \\ \boldsymbol{\dot{0}} \end{bmatrix} = \begin{bmatrix} \boldsymbol{G}_{ad}  &  \boldsymbol{G}_{af} \\ \boldsymbol{G}_{cd} &  \boldsymbol{G}_{cf}  \end{bmatrix} \begin{bmatrix} \boldsymbol{\dot{x}}_{d} \\ \boldsymbol{\dot{x}}_{f}\end{bmatrix}    \label{eq:reshuffled}                                                              
\end{equation}
where, 

\begin{equation*}
	\begin{aligned}
		  \boldsymbol{G}_{ad}  &= \begin{bmatrix} \boldsymbol{G}_{a1vx} &  \boldsymbol{G}_{a1vy} &  \boldsymbol{G}_{a1wz} \\ \boldsymbol{G}_{a2vx} &  \boldsymbol{G}_{a2vy} &  \boldsymbol{G}_{a2wz} \\  \boldsymbol{G}_{a3vx} &  \boldsymbol{G}_{a3vy} &  \boldsymbol{G}_{a3wz} \end{bmatrix}  \\    
		  \boldsymbol{G}_{af}  &= \begin{bmatrix} \boldsymbol{G}_{a1vz} &  \boldsymbol{G}_{a1wx} &  \boldsymbol{G}_{a1wy} \\ \boldsymbol{G}_{a2vz} &  \boldsymbol{G}_{a2wx} &  \boldsymbol{G}_{a2wy} \\  \boldsymbol{G}_{a3vz} &  \boldsymbol{G}_{a3wx} &  \boldsymbol{G}_{a3wy} \end{bmatrix}  \\ 
		  \boldsymbol{G}_{cd}  &= \begin{bmatrix} \boldsymbol{G}_{c1vx} &  \boldsymbol{G}_{c1vy} &  \boldsymbol{G}_{c1wz} \\ \boldsymbol{G}_{c2vx} &  \boldsymbol{G}_{c2vy} &  \boldsymbol{G}_{c2wz} \\  \boldsymbol{G}_{a3vx} &  \boldsymbol{G}_{c3vy} &  \boldsymbol{G}_{c3wz} \end{bmatrix}  \\ 
		  \boldsymbol{G}_{cf}  &= \begin{bmatrix} \boldsymbol{G}_{c1vz} &  \boldsymbol{G}_{c1wx} &  \boldsymbol{G}_{c1wy} \\ \boldsymbol{G}_{c2vz} &  \boldsymbol{G}_{c2wx} &  \boldsymbol{G}_{c2wy} \\  \boldsymbol{G}_{c3vz} &  \boldsymbol{G}_{c3wx} &  \boldsymbol{G}_{c3wy} \end{bmatrix},  \\ 
		  \boldsymbol{\dot{x}}_{d} = \begin{bmatrix} v_x \\ v_y \\ \omega_z \end{bmatrix}, \boldsymbol{\dot{x}}_{f} = \begin{bmatrix} v_z \\ \omega_x \\ \omega_y \end{bmatrix}
	\end{aligned} \\ 
\end{equation*}

Likewise, the stiffness matrix can be reorganized following the order of task space motion in Eq. (\ref{eq:reshuffled}).   

\section{Stiffness Formulation} \label{sec: stiffness_formulation}
 
Stiffness formulation in parallel robots designed for machining purpose is essential for determining their performance, precision, and load capacity. It involves calculating a stiffness matrix that quantifies the relationship between applied forces and the resulting displacements, crucial for maintaining accuracy and reducing vibrations during operation. This matrix is derived through methods such as the Jacobian matrix analysis, the virtual work principle, energy methods, and finite element analysis (FEA). Adequate stiffness ensures the machine's stability, responsiveness, and operational integrity across various applications, making it a fundamental aspect of PKM design and analysis. This section evaluates the stiffness of Sprint Z3 machine. 

 \subsection{Stiffness modeling and analysis}
 
 It is crucial to understand that the constraint space also has important roles in stiffness of the structure as it does in the kinematics. These manipulators cannot generate forces in all six degrees of freedom, leading to constraints that, while unable to produce work, must be included in stiffness analysis. Therefore, this section develops the stiffness models for the Sprint Z3, taking into account all significant component compliances and structural constraints. 
 
 For simplicity,  the following assumptions hold in this paper. 
 
 \texttt{Assumption one:} Fixed and moving plates are rigid and these parts are not considered as a compliant body. 
 
 It is important to note that this is a static analysis, which disregards any inertia-induced bending in the direction of freedom.
 Considering each limb components as a series of connected springs, the moving plate deflection can be expressed with the following relation.
 
 \begin{equation}
 	\delta\boldsymbol{x} =\sum_{j_a=1}^{f}\$_{ai}\delta{q}_{ai} +\sum_{j_c=1}^{6-f}\$_{ci}\delta{q}_{ci}
 	\label{one}
 \end{equation}
 In Eq. (\ref{one}), $\delta{q}_{ai}$ and $\delta{q}_{ci}$ are the intensity of corresponding joint twist or can be called the linear/angular deflection values in the respective directions. Spatial vectors $\$_{ai}$ and $\$_{ci}$ corresponding to the elastic deflection screws. This relation implies the joint and moving platform infinitesimal displacements can be mapped by using the limb Jacobian. Consequently, their inverse relation is obtained as: 
 
 \begin{equation}
 	\begin{aligned}
 		\delta{\boldsymbol{q}}                                                                                    &= \boldsymbol{G}^\top \delta{\boldsymbol{\mathcal{X}}} \\ 
 		\begin{bmatrix} \delta{\boldsymbol{q}}_a \\ \delta{\boldsymbol{q}}_c \end{bmatrix}                       &= \begin{bmatrix}  \boldsymbol{G}_{av}^{\top} & \boldsymbol{G}_{aw}^{\top} \\  \boldsymbol{G}_{aw}^{\top} & \boldsymbol{G}_{cw}^{\top} \end{bmatrix} \begin{bmatrix} \delta\boldsymbol{r} \\ \delta\boldsymbol{\alpha} \end{bmatrix} 
 	\end{aligned}  \label{eq:stif_irk}
 \end{equation}


 Lte $\boldsymbol{\tau}=\begin{bmatrix}
 	\boldsymbol{f}^\top&\boldsymbol{m}^\top
 \end{bmatrix}^\top$ be applied on the moving platform and $\delta\boldsymbol{x}=\begin{bmatrix}
 	\delta\boldsymbol{r}^\top & \delta \boldsymbol{\alpha}^\top
 \end{bmatrix}^\top$ be deflection induced by $\boldsymbol{\tau}$ where, $\delta\boldsymbol{r}$ and  $\delta \boldsymbol{\alpha}$ are linear and angular deflections. The manipulator in a state of equilibrium under the action of external forces (and moments),$\boldsymbol{\tau}$, and internal reaction forces $\boldsymbol{w}$, imposes the net work done on the system to be zero for any arbitrary small, virtual displacement. This condition ensures that the system's internal energy is conserved, and it does not spontaneously move or deform further, which would imply an imbalance. This phenomenon is described by the virtual work principle (vwp) as follows.
 
 \begin{equation}
 	\delta{\boldsymbol{\mathcal{X}}^\top \boldsymbol{\tau} = \delta{\boldsymbol{q}}^\top \boldsymbol{w} \label{eq:vrp_generic}
 } \end{equation}
 
 where $ \boldsymbol{w} $ is the reaction force which counteract the applied external wrench to maintain equilibrium. $ \delta{\boldsymbol{\mathcal{X}}} $ and $ \delta{\boldsymbol{q}} $ are infinitesimally small displacements produces by $ \boldsymbol{\tau} $ and $ \boldsymbol{w} $ that do not alter the external forces acting on the system. From Eq.(\ref{eq:irk}) and Eq. (\ref{eq:Gprs}), we derive Jacobian matrices that maps those deflections, encompassing the constraints. Hence, Eq. (\ref{eq:vrp_generic}) can be rewritten as: 
 
 \begin{equation}
 	\begin{aligned}
 		\delta{\boldsymbol{\mathcal{X}}}^\top \boldsymbol{\tau} = \delta{\boldsymbol{\mathcal{X}}}^\top \boldsymbol{G}   \boldsymbol{w} \\ 
 	\end{aligned} \label{eq:vrp_gen_exp}
 \end{equation}
 
 \texttt{Assumption two:} In the analysis, we model the individual components within a limb as a series of linearly connected elements or spring systems. Therefore, the axial and bending compliance of joints, limbs, lead screws and their associated elements are taken into account as a compliant element.  By adhering to Hooke's Law, which posits that the force required to extend or compress a spring by a certain distance is directly proportional to that distance, we formulate this relationship as:
 
 \begin{equation}
 	\boldsymbol{w} = \boldsymbol{\mathcal{K}} \delta{\boldsymbol{q}} \label{eq:hooks_law}
 \end{equation}
 where $\boldsymbol{\mathcal{K}} \in \mathbb{R}^{6 \times 6}$ is the actuation and constraint stiffness matrix, $\boldsymbol{w} \in \mathbb{R}^6$ is the force applied, and $\delta{\boldsymbol{q}} \in \mathbb{R}^6$ is the displacement caused by the force. This formulation underpins our assumption of linear elasticity for the components, facilitating the analysis of their response to applied forces.  
 
 Reorganizing Eq. (\ref{eq:vrp_gen_exp}) we get: 
 
 \begin{equation}
 	\boldsymbol{\tau} = \boldsymbol{G}  \boldsymbol{w}  \label{eq:tau_init}
 \end{equation}
 
 Then, by substituting Eq. (\ref{eq:stif_irk}) and Eq. (\ref{eq:hooks_law}) into Eq. (\ref{eq:tau_init}), we obtain: 
 
 \begin{equation}
 	\begin{aligned}
 		\boldsymbol{\tau} &= \boldsymbol{G} \boldsymbol{\mathcal{K}} \delta{\boldsymbol{q}} \\ 
 		&= \boldsymbol{G} \boldsymbol{\mathcal{K}} \boldsymbol{G}^\top \delta{\boldsymbol{\mathcal{X}}} \\ 
 		&= \boldsymbol{K}\delta{\boldsymbol{\mathcal{X}}}
 	\end{aligned} \label{eq:tau}
 \end{equation}
 
 Equation (\ref{eq:tau}) can be further expanded for unveiling the details as: 
 
 \begin{equation}
 	\begin{bmatrix} \boldsymbol{\tau}_a \\ \boldsymbol{\tau}_c	\end{bmatrix}  = \begin{bmatrix} \boldsymbol{K}_{a\parallel} & \boldsymbol{K}_{a\perp}	 \\ \boldsymbol{K}_{c\parallel} & \boldsymbol{K}_{c\perp} \end{bmatrix} \begin{bmatrix} \delta{\boldsymbol{r}} \\ \delta{\boldsymbol{\alpha}}	\end{bmatrix} 
 \end{equation}
 
 \section{Formulation of stiffness matrix} \label{sec: matrix_k}
 
\begin{figure}[htb!]
 	\centering
 	\includegraphics[width=0.7\columnwidth]{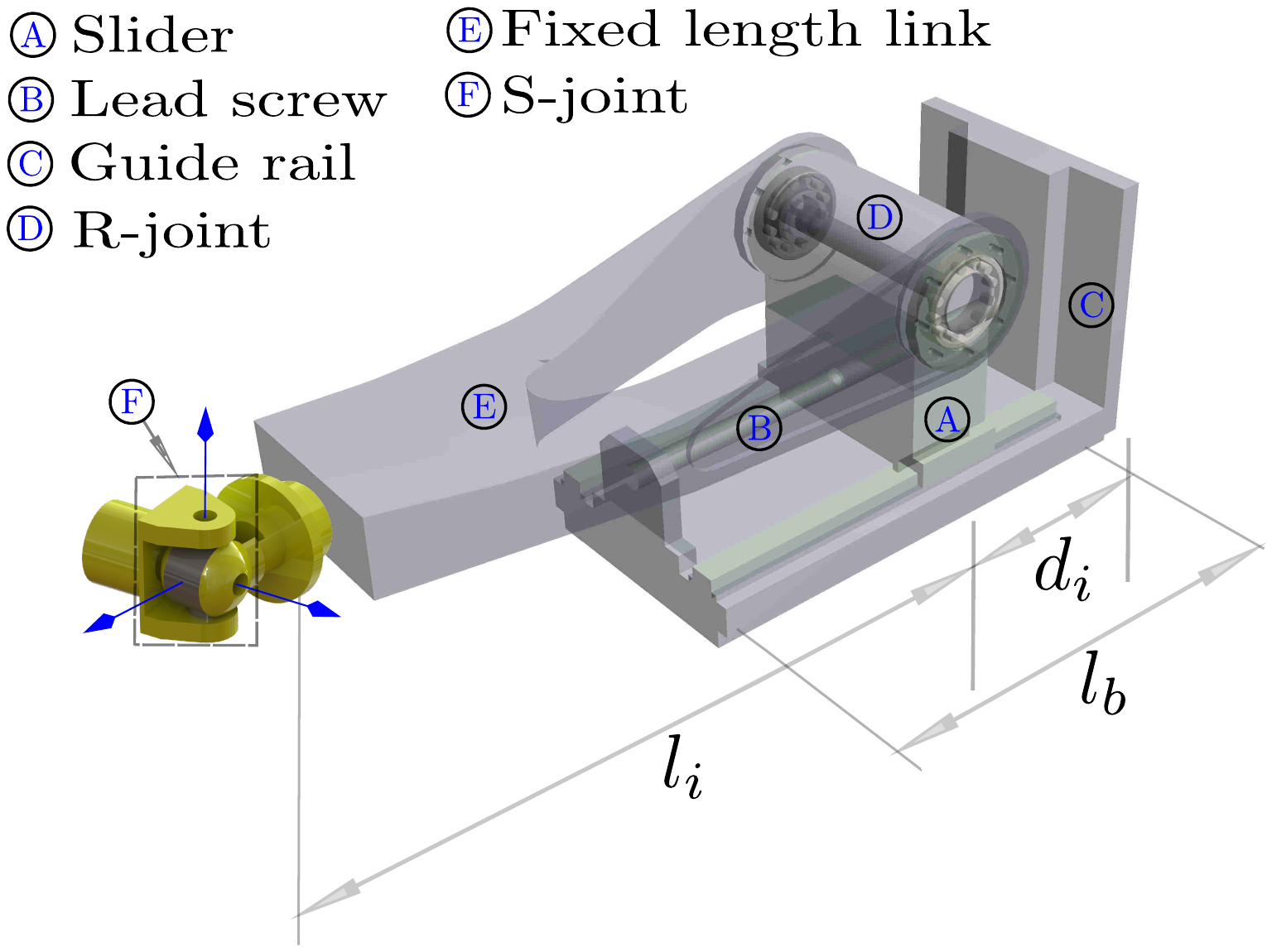} 
 	\caption{Sprint Z3 Limb}
 	\label{fig:limb}
 \end{figure}

 The formulation of the stiffness matrix $\boldsymbol{\mathcal{K}}$ entails calculating the axial and torsional stiffness of the acting components related to actuation and constraints and compiling their elasticity information into a matrix form. For convenience, the elastic elements of the manipulator are categorized into three main components,i.e., 1)  Carriage assembly ($CA$), which includes the slider, lead screw, and guide rail. 2) Revolute joints ($RJ$) and 3) Limb body ($LB$) comprising a fixed-length link and spherical joints (Fig. \ref{fig:limb}). 
 
 This grouping not only simplifies the analysis but also identifies sub-systems that could be independently targeted for design enhancements.
 
\texttt{Assumption three:} All joints are assumed to be friction-less.  
  
 The axial stiffness for each limb is calculated using the equation:
 
 \begin{equation}
 	\dfrac{1}{k_{ai}}  = \dfrac{1}{k_{cai}} + \dfrac{1}{k_{rji}} + \dfrac{1}{k_{lbi}} \label{eq:axial_stiffness}
 \end{equation}

where:
\begin{itemize}
	\item $k_{cai}, k_{rji},$ and $k_{lbi}$ denote the axial stiffness coefficients of the Carriage assembly, Revolute joints, and Limb body, respectively.
	\item $k_{ai}$ represents the overall axial stiffness coefficient for limb $i$.
\end{itemize}

The torsional stiffness coefficients at the distal point of each limb are measured along the axis of the revolute joint, which is parallel to $\boldsymbol{s}_{2i\parallel}$. The torsional stiffness, $k_{ci}$, can be determined by calculating the inverse sum of its coefficients of the spherical joint ($S$ joint) and the fixed length link.

\begin{equation}
	\dfrac{1}{k_{ci}} = 	\dfrac{1}{k_{si}} + 	\dfrac{1}{k_{fli}}
\end{equation}
where $k_{si}$ is configuration dependent stiffness coefficient of the spherical joint along and about the corresponding axes, while fixed-length-link stiffness coefficient $k_{fli}$ is obtained via FEM analysis software. The derivation of  $k_{si}$ is given as follows. As the spherical joint is the product of three revolute joints, it can be decomposed to three 1 DoF joints. The corresponding stiffness matrix of the spherical joint is then obtained as  

\begin{equation}
	 \boldsymbol{\mathcal{K}}_s = \boldsymbol{R}_\mathcal{F}^\top  \boldsymbol{\bar{\mathcal{K}}}_s \boldsymbol{R}_\mathcal{F}  
\end{equation}

where $\boldsymbol{R}_\mathcal{F} = \begin{bmatrix} \boldsymbol{x} & \boldsymbol{y} & \boldsymbol{z}	\end{bmatrix} $ is the rotation matrix of spherical joint frame with respect the fixed frame and $\boldsymbol{\bar{\mathcal{K}}}_s = \begin{bmatrix}  k_{six} & 0 & 0 \\ 0 &  k_{siy} & 0 \\ 0 & 0 &  k_{siz}	\end{bmatrix}   $ is a set of stiffness coefficients for the spherical joints corresponding to the three axes. This approach effectively combines the contributions of both elements to the overall torsional stiffness of each limb. The orientation matrix can $\boldsymbol{R}_\mathcal{F}$ can be obtained from the following relationship: 

 \begin{equation}
	\begin{aligned}
		\boldsymbol{H}_p &= \boldsymbol{H}\boldsymbol{R}_z(\xi_i)\boldsymbol{H}\boldsymbol{t}_x(r_b)\boldsymbol{H}\boldsymbol{t}_z(d_i)\boldsymbol{H}\boldsymbol{R}_y(\theta_{2i})\boldsymbol{H}\boldsymbol{T}_z(l_i)\times \\& ~~~~\boldsymbol{H}\boldsymbol{R}_y(\theta_{3i})\boldsymbol{H}\boldsymbol{R}_x(\theta_{4i})\boldsymbol{H}\boldsymbol{R}_z(\theta_{5i})  \\ 
		&=  \boldsymbol{H}_{pr}\boldsymbol{H}_s
	\end{aligned} \label{eq:limb_trans}
\end{equation}
where 

$\boldsymbol{H}_{pr} = \boldsymbol{H}\boldsymbol{R}_z(\xi_i)\boldsymbol{H}\boldsymbol{t}_x(r_b)\boldsymbol{H}\boldsymbol{t}_z(d_i)\boldsymbol{H}\boldsymbol{R}_y(\theta_{2i})\boldsymbol{H}\boldsymbol{T}_z(l_i)$, $\boldsymbol{H}_s = \boldsymbol{H}\boldsymbol{R}_y(\theta_{3i})\boldsymbol{H}\boldsymbol{R}_x(\theta_{4i})\boldsymbol{H}\boldsymbol{R}_z(\theta_{5i})$ and $\boldsymbol{H} \in \mathbb{R}^{4\times 4}$ is a homogeneous transformation matrix. 
In Equation \ref{eq:limb_trans}, $\boldsymbol{H}_p = \begin{bmatrix} \boldsymbol{R} & \boldsymbol{p} \\ \boldsymbol{0} & 1 \end{bmatrix}  $ represents the pose of the moving platform. $\boldsymbol{H}{pr}$ and $\boldsymbol{H}s$ denotes the information for the prismatic + revolute and spherical joints, respectively, as derived from the limb configuration.
Then, the spherical joint angles are simply obtained by extracting the rotational terms and equating the left-hand side and right-hand side of $\boldsymbol{H}{pr}^{-1}\boldsymbol{H}_p = \boldsymbol{H}_s$. Then, $\boldsymbol{R}_\mathcal{F}$ is established out of it. 

Referring to Fig. (\ref{fig:limb}), the carriage assembly rigidity can be modeled as a linear function of length $d_i$ while the fixed length link stiffness coefficient (${k}_{fl}$) is constant. 

\begin{equation}
	\begin{split}
		{k}_{ls}^{-1}=(d_i)/EA_{ls} + k_{gr}^{-1} + k_{sl} \\ 
		{k}_{fl}^{-1} = L/EA_l
	\end{split}
\end{equation} 
where  $EA_{ls}$ and $EA_l$ are the Young's moduli of the lead screw and the fixed-length link, respectively, and $k_{gr}$ and $k_{sl}$ are the stiffness coefficients of the guide-rail and slider.

\begin{table}[h]
	\centering
	\caption{Axial Stiffness coefficients for Components of Sprint Z3}
	\label{tab:axial_stiffness_values_sprint_z3}
	\begin{tabular}{l c}
		\hline
		Component & Axial Stiffness ($k_a$, in $N/m$) \\
		\hline
		Carriage Assembly ($CA$) & $3.8 \times 10^7$ \\
		Revolute Joints ($RJ$) & $3.2 \times 10^9$ \\
		Limb Body ($LB$) & $976 \times 10^6$ \\
		\hline
	\end{tabular}
\end{table}

\begin{table}[h]
	\centering
	\caption{Torsional Stiffness coefficients for Components of Sprint Z3}
	\label{tab:detailed_torsional_stiffness_sprint_z3}
	\begin{tabular}{l c}
		\hline
		Component & Torsional Stiffness ($k_c$, in $Nm/deg$) \\
		\hline
		Spherical Joint ($S$) & $kcis = 8.9 \times 10^5 $ \\
		Limb Body Assembly ($LB$) & $kcib = 7.8  \times 10^5 $ \\
		\hline
	\end{tabular}
\end{table}

The comprehensive stiffness matrix $\boldsymbol{\mathcal{K}}$ is an upper diagonal matrix formulated by assembling the stiffness contributions from all limb components into a diagonal matrix $\boldsymbol{\mathcal{K}}_{ac}$ and integrating coordinate transformation and Jacobian matrices adjustments:

\begin{equation}
	\begin{aligned}
		\boldsymbol{K} &= \boldsymbol{G} \boldsymbol{\mathcal{K}}\boldsymbol{G}^\top \\
		&= \begin{bmatrix} \boldsymbol{G}_a & \boldsymbol{G}_c \end{bmatrix}  \begin{bmatrix}  \boldsymbol{\mathcal{K}}_a & \boldsymbol{0} \\ \boldsymbol{0} & \boldsymbol{\mathcal{K}}_c \end{bmatrix}  \begin{bmatrix}  \boldsymbol{G}_a^\top  \\ \boldsymbol{G}_c^\top \end{bmatrix} 
	\end{aligned}
\end{equation}
where $\boldsymbol{G}$ embodies the combined actuation ($\boldsymbol{G}_a$) and constraint ($\boldsymbol{G}_c$) Jacobian matrices, and if the target coordinate is not the center of the moving platform, it can be adjusted through the $ 6 \times 6$ adjoint transformation matrix. 

Furthermore, the bending stiffness is included in $\boldsymbol{\mathcal{K}}$ by evaluating the reciprocal sum of the stiffness coefficients for each component, ensuring a comprehensive representation of the manipulator's elastic response to external forces and maintaining structural integrity under load.

\section{Numerical Simulation} \label{sec:numerical_simulation}
Here, the numerical simulation is performed with the parameters provided below to visualize the stiffness value distribution over the entire workspace.  
\begin{table}[!htb]
	\centering
	\caption{Geometric Parameters of the  Mechanism}
	\label{tab:geometric_parameters_rps}
	\begin{tabular}{lcc}
		\hline
		{Parameter} &  {Description} &  {Value} \\
		\hline
		$rb$ & Radius of the base & 326.923 mm \\
		$ra$ & Radius of moving plate & 250.00 mm \\
		$\theta = \psi$ &Orientation angle & $\pm 40^o$ \\ 
		\hline
	\end{tabular}
\end{table}

where $l$ in the case of RPS manipulator is the prismatic joint length which will be extended and shortened based on the orientation.

\begin{figure}[!htb]
	\centering
	\begin{subfigure}{0.5\columnwidth}
		\centering
		\includegraphics[width=\textwidth]{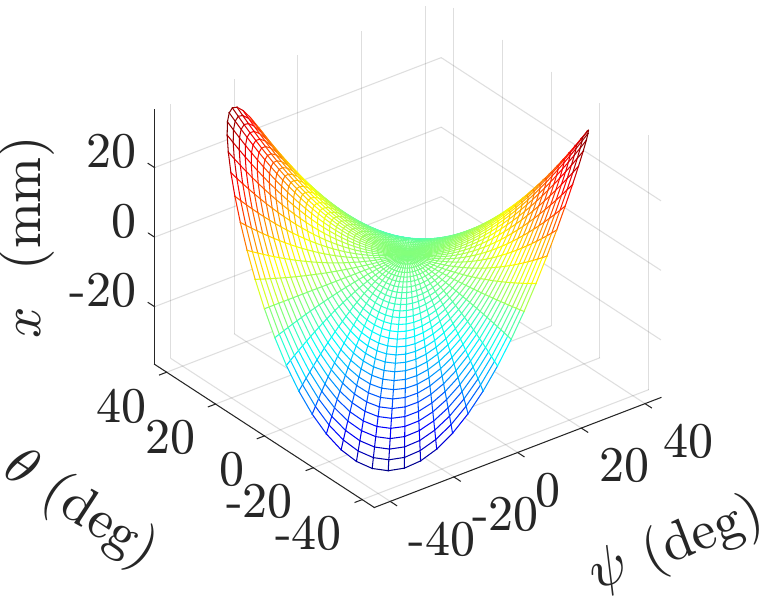}
		\caption{}\label{fig:x_par}
	\end{subfigure}\hfill
	\begin{subfigure}{0.5\columnwidth}
		\centering
		\includegraphics[width=\textwidth]{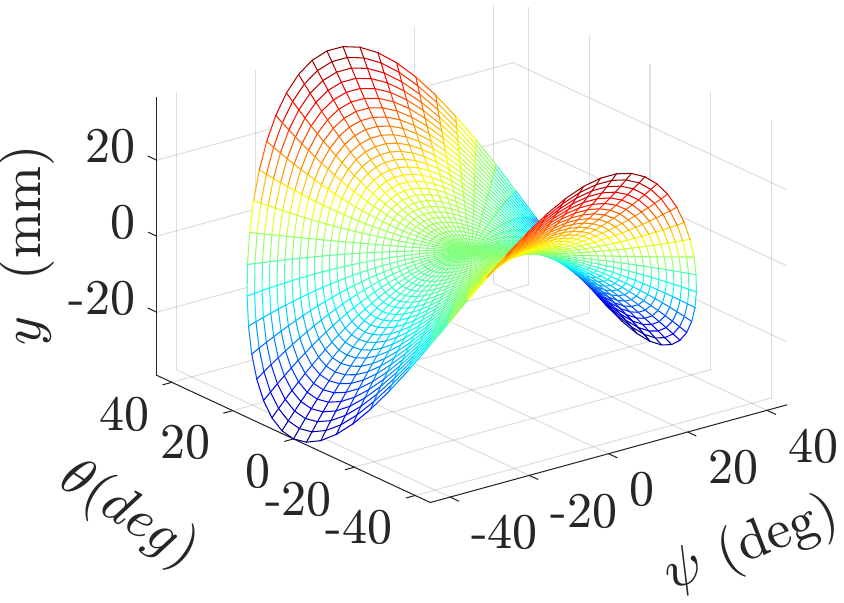}
		\caption{}\label{fig:y_par}
	\end{subfigure}\hfill
	\caption{The parasitic motion of the manipulator.  (a) $x$ and (b) $y$ axes translation }
	\label{fig:par}
\end{figure}

\begin{figure}[!htb]
	\centering
	\begin{subfigure}{0.5\columnwidth}
		\centering
		\includegraphics[width=\textwidth]{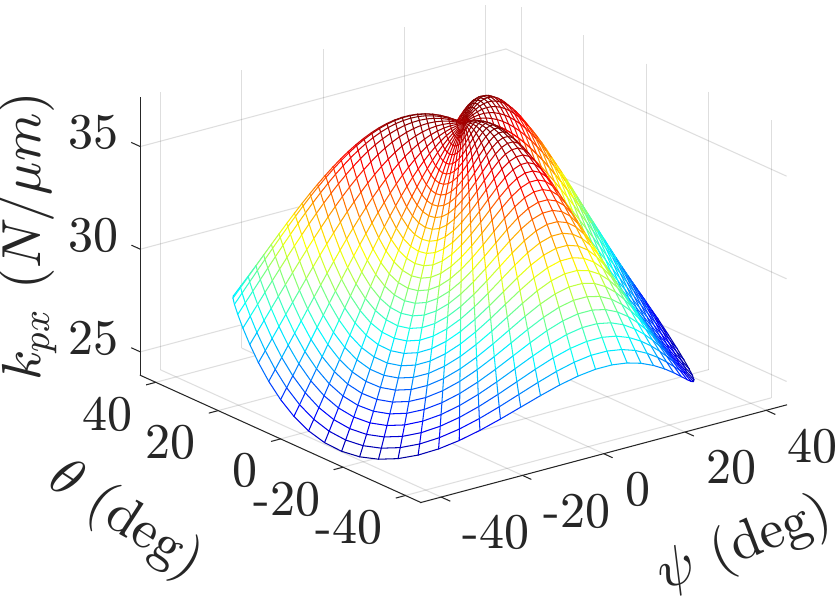}
		\caption{}\label{fig:kpx_z3}
	\end{subfigure}\hfill
	\begin{subfigure}{0.5\columnwidth}
		\centering
		\includegraphics[width=\textwidth]{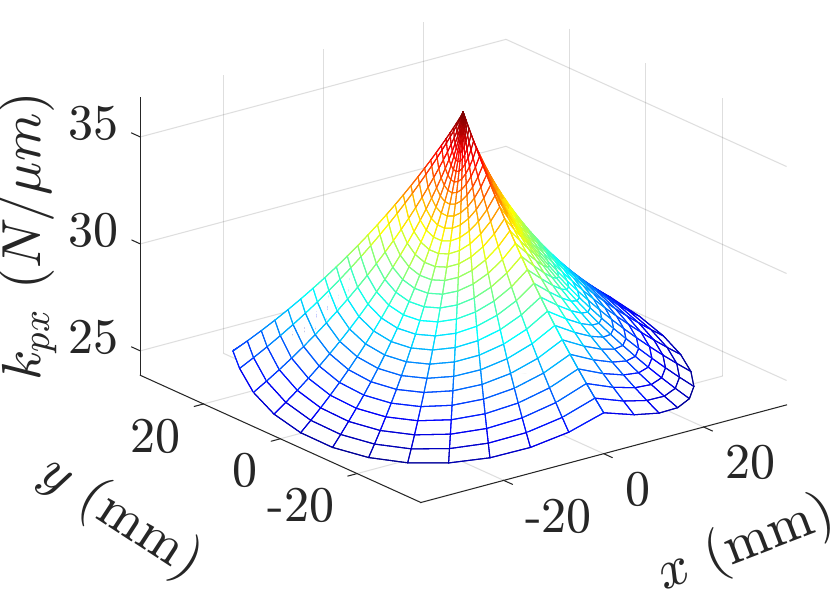}
		\caption{}\label{fig:kpx_z3_param}
	\end{subfigure}\hfill
	\caption{The x-axis axial stiffness distribution. (a) Across orientation space. (b) Across parasitic space}
	\label{fig:kpx}
\end{figure}

\begin{figure}[!htb]
	\centering
	\begin{subfigure}{0.5\columnwidth}
		\centering
		\includegraphics[width=\textwidth]{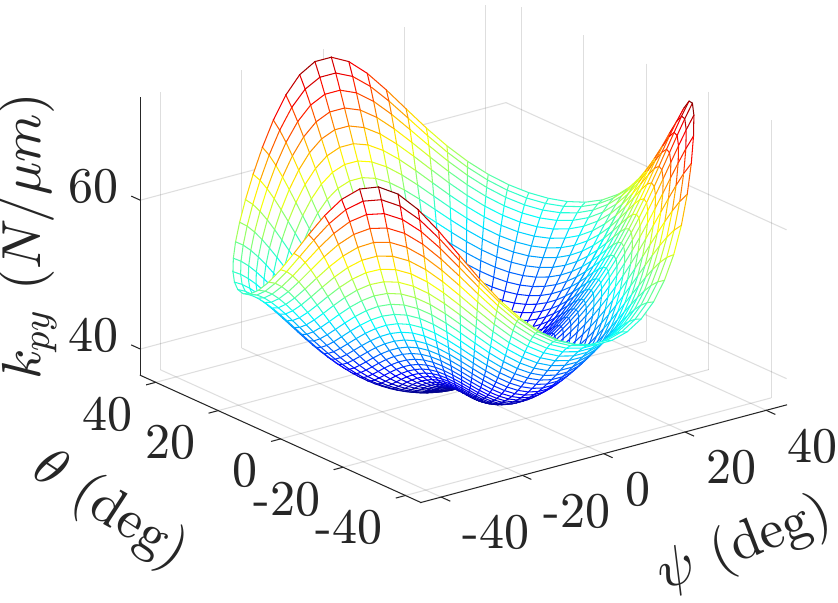}
		\caption{}\label{fig:kpy_z3}
	\end{subfigure}\hfill
	\begin{subfigure}{0.5\columnwidth}
		\centering
		\includegraphics[width=\textwidth]{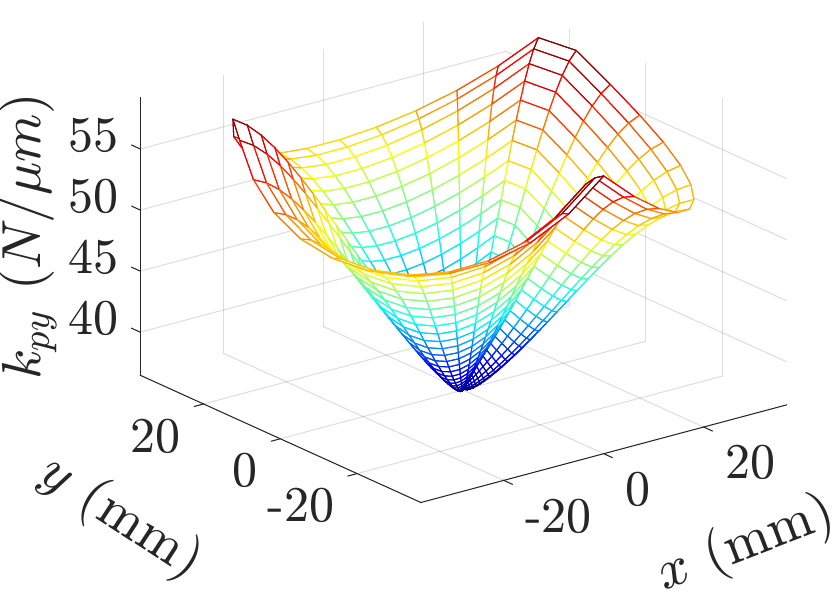}
		\caption{}\label{fig:kpy_z3_param}
	\end{subfigure}\hfill
	\caption{The y-axis axial stiffness distribution. (a) Across orientation space. (b) Across parasitic space}
	\label{fig:kpy}
\end{figure}

\begin{figure}[!htb]
	\centering
	\begin{subfigure}{0.5\columnwidth}
		\centering
		\includegraphics[width=\textwidth]{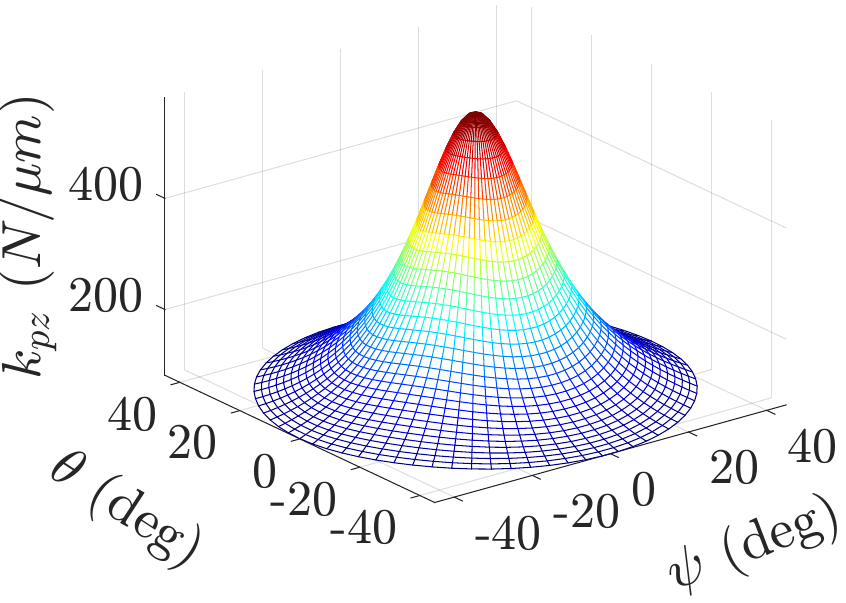}
		\caption{}\label{fig:kpz_z3}
	\end{subfigure}\hfill
	\begin{subfigure}{0.5\columnwidth}
		\centering
		\includegraphics[width=\textwidth]{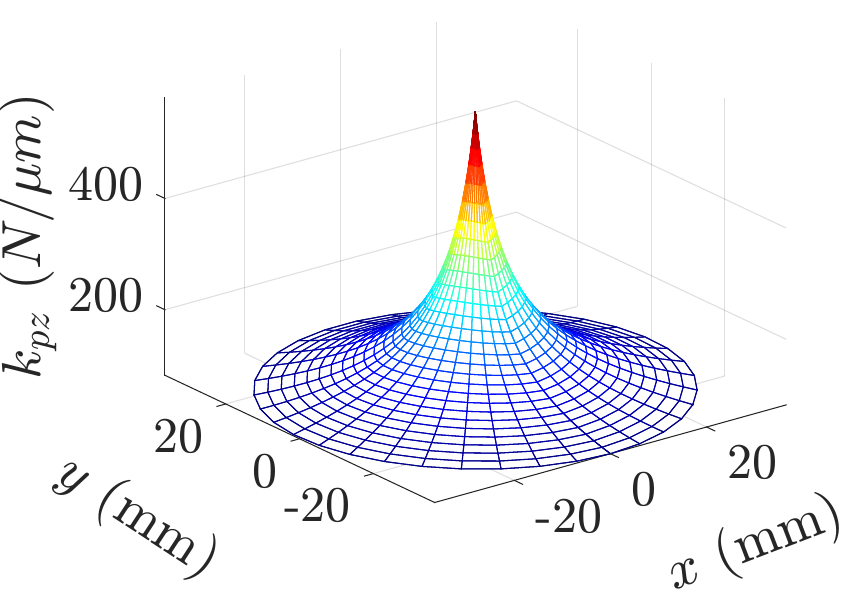}
		\caption{}\label{fig:kpz_z3_param}
	\end{subfigure}\hfill
	\caption{The z-axis axial stiffness distribution. (a) Across orientation space. (b) Across parasitic space}
	\label{fig:kpz}
\end{figure}

\begin{figure}[!htb]
	\centering
	\begin{subfigure}{0.5\columnwidth}
		\centering
		\includegraphics[width=\textwidth]{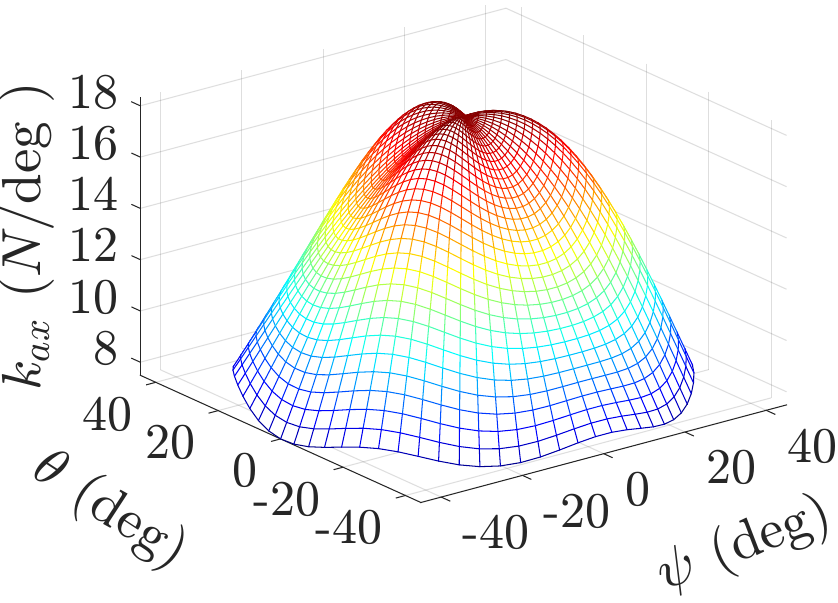}
		\caption{}\label{fig:kax_z3}
	\end{subfigure}\hfill
	\begin{subfigure}{0.5\columnwidth}
		\centering
		\includegraphics[width=\textwidth]{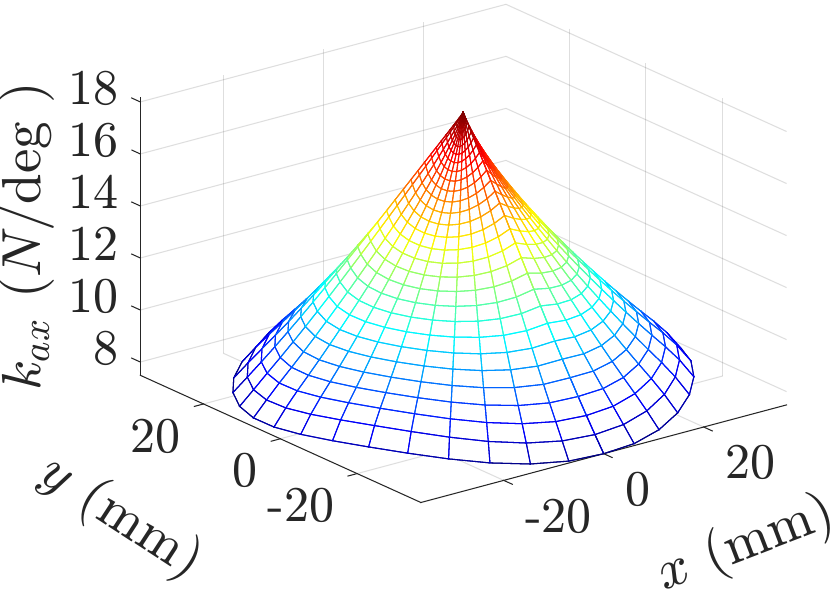}
		\caption{}\label{fig:kax_z3_param}
	\end{subfigure}\hfill
	\caption{The x-axis torsional stiffness distribution. (a) Across orientation space. (b) Across parasitic space}
	\label{fig:kax}
\end{figure}

\begin{figure}[!htb]
	\centering
	\begin{subfigure}{0.5\columnwidth}
		\centering
		\includegraphics[width=\textwidth]{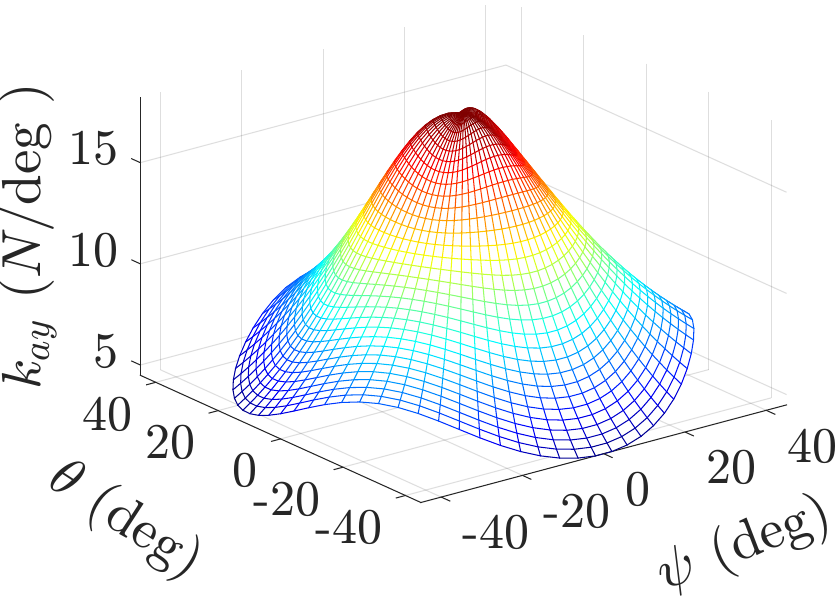}
		\caption{}\label{fig:kay_z3}
	\end{subfigure}\hfill
	\begin{subfigure}{0.5\columnwidth}
		\centering
		\includegraphics[width=\textwidth]{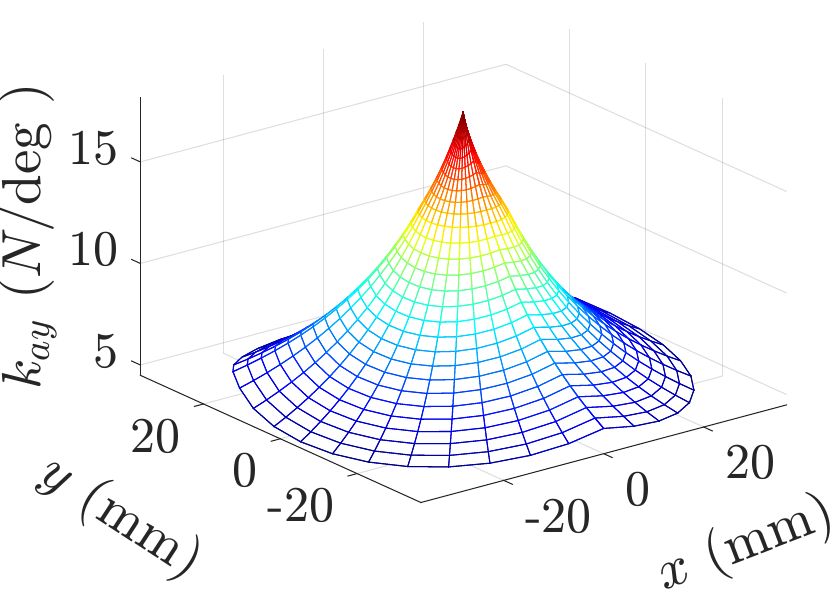}
		\caption{}\label{fig:kay_z3_param}
	\end{subfigure}\hfill
	\caption{The y-axis torsional stiffness distribution. (a) Across orientation space. (b) Across parasitic space}
	\label{fig:kay}
\end{figure}

\begin{figure}[!htb]
	\centering
	\begin{subfigure}{0.5\columnwidth}
		\centering
		\includegraphics[width=\textwidth]{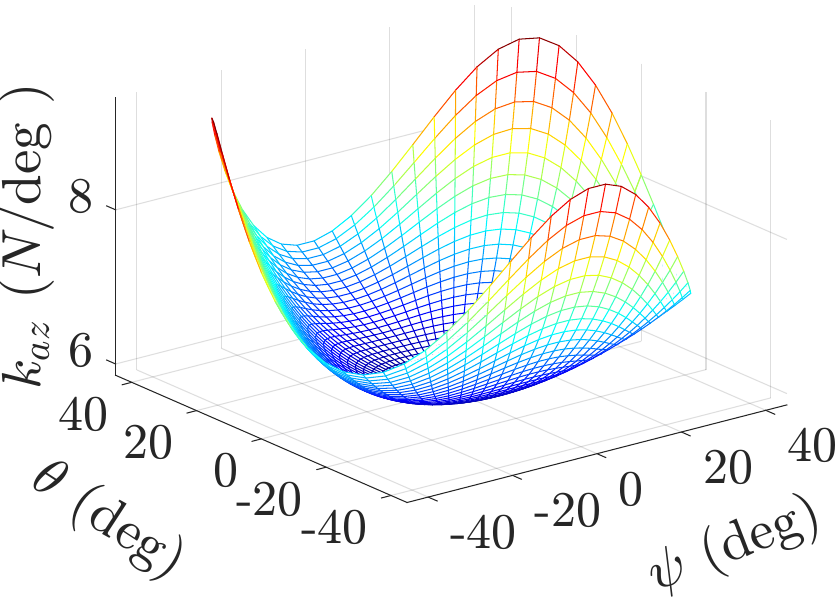}
		\caption{}\label{fig:kaz_z3}
	\end{subfigure}\hfill
	\begin{subfigure}{0.5\columnwidth}
		\centering
		\includegraphics[width=\textwidth]{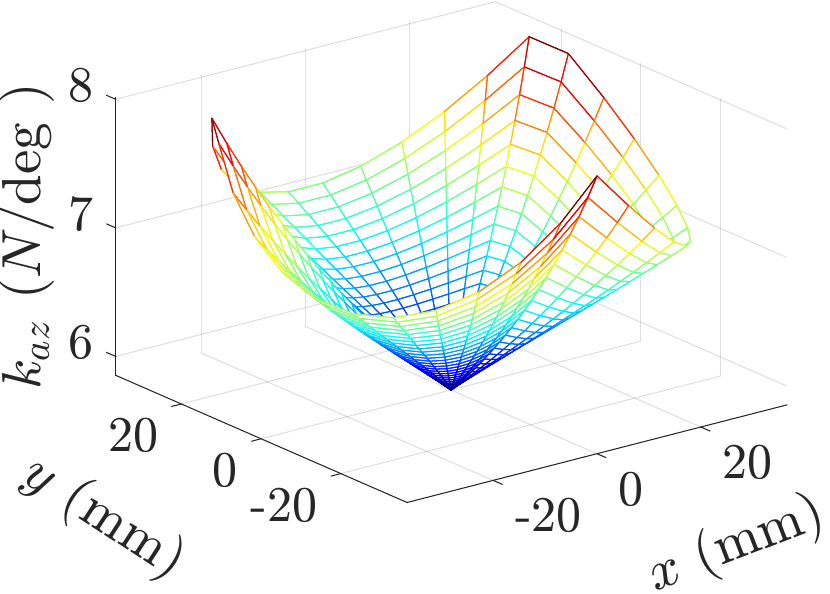}
		\caption{}\label{fig:kaz_z3_param}
	\end{subfigure}\hfill
	\caption{The z-axis torsional stiffness distribution. (a) Across orientation space. (b) Across parasitic space}
	\label{fig:kaz}
\end{figure}

In Fig. \ref{fig:par}, the x and y axes parasitic motion is shown at a particular height of the manipulator. This motion is automatically generated using Eq. (\ref{eq:projection}). The simulation results shown from Figs. \ref{fig:kpx} to \ref{fig:kaz} illustrate the axial and torsional stiffness distributions over the independent and parasitic motion spaces shown in Fig. \ref{fig:par}.  Stiffness variation across the orientation space, represented by Figs. \ref{fig:kpx_z3}, \ref{fig:kpy_z3}, \ref{fig:kpz_z3}, \ref{fig:kax_z3}, \ref{fig:kay_z3} and \ref{fig:kaz_z3}, typically peaks at, indicating maximum stiffness at the zero orientation. However, Figs. \ref{fig:kpy_z3} and \ref{fig:kaz_z3} depict minimum stiffness values for the y-axis and z-axis at this orientation. 
Likewise, the stiffness profile over the parasitic space, as shown in Figs. \ref{fig:kpx_z3_param}, \ref{fig:kpy_z3_param}, \ref{fig:kpz_z3_param}, \ref{fig:kax_z3_param}, \ref{fig:kay_z3_param} and \ref{fig:kaz_z3_param}, suggests stiffness over the uncontrolled translational space along the x and y axes, peaking at the center. 
From the results, we can observe that the stiffness over the rotational workspace is more uniform, whereas in the parasitic motion space, it covers a narrower range. Although the peak stiffness is consistent across all spaces, variations in the parasitic motion space configurations indicates the need for careful design consideration in robots with such an uncontrolled parasitic displacement.

\section{Conclusions} \label{sec:conclusions}
In this work, we address the limitations stemming from the overlooked consideration of stiffness evaluation in relation to parasitic motion space. The stiffness distribution of the Sprint Z3 across independent and parasitic motion spaces is evaluated by formulating a stiffness model in relation to the parasitic motion at the velocity level.  With parameters in Table \ref{tab:geometric_parameters_rps}, stiffness distribution across these spaces is evaluated. The simulation results reveal that higher stiffness values are more uniformly distributed across the orientation space than the parasitic motion space. This indicates that the lower stiffness profile over the undesired translational motion space could pose rigidity issues. Thus, special emphasis must be given on the stiffness capabilities of manipulators with parasitic motion to improve certain components.

\section*{Acknowledgment}

This research was supported by the Robotics Research Center of Yuyao (Grant No. KZ22308), National Natural Science Foundation of China under the Youth Program (Grant No. 509109-N72401) and the 2023 National High-level Talent Project, also within the Youth Program (Grant No. 588020-X42306/008).

\nocite{*}
\bibliographystyle{IEEEtran}
\bibliography{IEEEabrv,remar}

\end{document}